\documentclass[conference]{IEEEtran}
\IEEEoverridecommandlockouts
\usepackage{cite}
\usepackage{amsmath,amssymb,amsfonts}
\usepackage{algorithmic}
\usepackage{graphicx}
\usepackage{textcomp}
\usepackage{xcolor}
\usepackage{comment}
\usepackage{algorithm2e}
\usepackage{subcaption}
\def\BibTeX{{\rm B\kern-.05em{\sc i\kern-.025em b}\kern-.08em
    T\kern-.1667em\lower.7ex\hbox{E}\kern-.125emX}}
\begin{document}

\title{LegoNet: Memory Footprint Reduction Through
Block Weight Clustering\\
}

%\begin{comment}
\author{\IEEEauthorblockN{1\textsuperscript{st} Joseph Bingham}
\IEEEauthorblockA{\textit{Dept of Electrical and Computer Engineering} \\
\textit{Rutgers University}\\
New Brunswick, New Jersey \\
joseph.bingham@rutgers.edu}
\and

\IEEEauthorblockN{2\textsuperscript{nd} Noah Green}
\IEEEauthorblockA{\textit{Dept of Computer Science} \\
\textit{Rutgers University}\\
New Brunswick, New Jersey \\
ncg44@scarletmail.rutgers.edu}
\and

\IEEEauthorblockN{3\textsuperscript{rd} Saman Zonouz}
\IEEEauthorblockA{\textit{Dept of Electrical and Computer Engineering} \\
\textit{Rutgers University}\\
New Brunswick, New Jersey \\
saman.zonouz@rutgers.edu}
}
%\end{comment}

%\author{\IEEEauthorblockN{1\textsuperscript{st} Anonymus}
%\IEEEauthorblockA{\textit{Department of Anonymus} \\
%\textit{Anonymus University}\\
%Anonymus, AN \\
%Anon@Anonymus.edu}
%}

\makeatletter

\def\ps@IEEEtitlepagestyle{%
  \def\@oddfoot{\mycopyrightnotice}%
  \def\@evenfoot{}%
}
\def\mycopyrightnotice{%
  {\footnotesize 978-1-6654-6297-6/22/\$31.00 ©2022 IEEE\hfill}% <--- Change here
  \gdef\mycopyrightnotice{}% just in case
}

\maketitle
\begin{abstract}
As the need for neural network-based applications to become more accurate and powerful grows, so too does their size and memory footprint. With embedded devices, whose cache and RAM are limited, this growth hinders their ability to leverage state-of-the-art neural network architectures. In this work, we propose \textbf{LegoNet}, a compression technique that \textbf{constructs blocks of weights of the entire model regardless of layer type} and clusters these induced blocks. Using blocks instead of individual values to cluster the weights, we were able to compress ResNet-50 trained for Cifar-10 and ImageNet with only 32 4x4 blocks, compressing the memory footprint by over a factor of \textbf{64x without having to remove any weights} or changing the architecture and \textbf{no loss to accuracy}, nor retraining or any data, and show how to find an arrangement of 16 4x4 blocks that gives a compression ratio of \textbf{128x with less than 3\% accuracy loss}. This was all achieved with \textbf{no need for (re)training or fine-tuning}.

\end{abstract}
\noindent\textbf{Compression, Weight Clustering, Compact Representation}

\section{Introduction}
\label{introduction}

Deep neural networks (DNNs)~\cite{liu2017survey} have become a powerful tool in a wide range of critical domains such as healthcare, agriculture, and finance. As these algorithms get applied into more complex problems, their architectures have been increased in size as well~\cite{nguyen2021wide}. Due to their complexity to train, this has led to another phenomena where trained models are becoming more available for application designers to use, such as trained ResNet
%~\cite{DBLP:journals/corr/HeZRS15} 
and VGG
%~\cite{DBLP:journals/corr/SimonyanZ14a} 
on different datasets built for Keras.
%~\cite{chollet2015keras} 
and Pytorch. %~\cite{NEURIPS2019_9015}. 

These trained models have been used to by application designers to accelerate their production to great effect. However, for embedded applications and domains, users desire to put these larger models onto smaller and memory-limited devices. A prime example of such an application would be a mobile phone based application and in micro-controllers. In order to fit a state-of-the-art size model, such as VGG or ResNet, on these devices they must be compressed in order to reduce their memory footprint. This work proposes a method that allows for large models to execute on resource-restricted devices by utilizing an architecture-agnostic, block weight clustering technique. 
\begin{figure*}
    \centering
    \includegraphics[width=\textwidth]{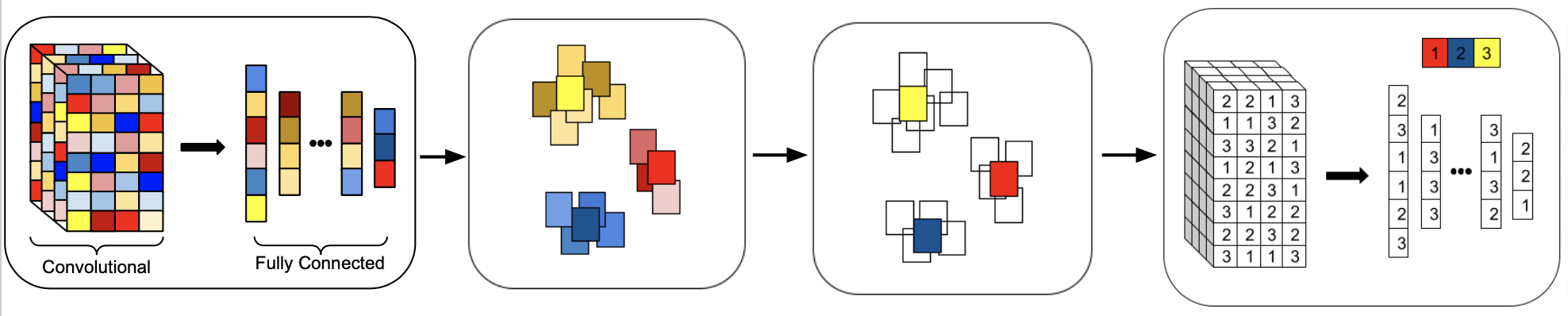}
    \caption{An overview of LegoNet algorithm. First a trained model's weight matrices are chunked into blocks, or Legos. Then the Legos are clustered into groups. Then the blocks are replaced by the index of the cluster they belong to. At inference time, the centroid representative of this cluster is used as the weight values.}
    \label{fig:legonet}
\end{figure*}

Current approaches for reducing model footprint, such as pruning~\cite{krishnan2019structural}, require fine-tuning, change the architecture of the model which reduces the fundamental capacity of the model as well as make them more difficult to integrate with off-the-shelf model pipelines that are relying on a certain architecture. Other methods, like knowledge distillation~\cite{Gou_2021} require training, which can be impractical if there is no data available, which would be the case if the user hopes to employ an off-the-shelf model. The current best techniques in weight-sharing~\cite{pmlr-v80-wu18h} only cluster based on single elements or based on subsections of rows of matrices. Further, they only cluster weights within a fixed context based on retraining. These contribute to lower compression ratios and need labelled data. 

An optimal solution to compressing neural networks for an embedded device would thus have the following qualities: 
\begin{itemize}
		\item{Efficient} A solution should compress the model's memory footprint with little to no affects on accuracy.
		\item{General} A solution should not depend on data or architectural features. It should take a large, trained model and compress it without needing retraining of any kind.  
		\item{Model Stable} A solution should not change the architecture of the model, that is to say the number of parameters and architecture should not change. This allows for off-the-self trained models to be used on embedded systems effortlessly. Further, it should not require any training or retraining. 
\end{itemize}		

In this paper, we propose \textbf{LegoNet}, a block-based weight clustering algorithm that can reduce the memory footprint by far more than the current state-of-the-art~\ref{tab:cr_comparisons} without removing any weights or changing the architecture (see Fig.~\ref{fig:legonet}~\footnote{This work was assisted by undergraduate Gilad Schneider.}), achieving as high as \textbf{128x compression} on ResNet-50 with less than \textbf{3\% accuracy loss} on ImageNet. We define a block to be a group of adjacent weights within a layer. As will be explained further in Section~\ref{methods}, since it is clustering on the weight matrices regardless of what type of layer they are in or which layer, LegoNet is data and architecture independent. 

The contributions of this work are:

\begin{enumerate}
		\item We describe the constraints present in embedded devices and the complications of trying to use a full trained model with them.
		\item We propose LegoNet as a solution to minimize the model footprint without the need for model retraining and provide a theoretical analysis thereof. 
		\item We validate LegoNet and compare its performance with the current state-of-the-art compression algorithms on known models and datasets. Code for reproducing our results using ResNet-50 on the ImageNet dateset will be provided in the supplementary material.  
		\item We present the \textbf{lossless} results, using LegoNet-A\footnote{LegoNet-A, or LegoNet-Accuracy, is the base LegoNet algorithm focused on accuracy, achieving no loss to the original models accuracy.}, as well as a method for finding the best number of clusters given a loss tolerance, using LegoNet-C\footnote{LegoNet-C, or LegoNet-Compression, is the base LegoNet algorithm focused on compression given an error tolerance.}, which achieves \textbf{128x compression ratio} on the 2012 ImageNet dataset with \textbf{less than \%3 loss} to accuracy.
\end{enumerate}

\begin{comment}
\paragraph{How LegoNet is different} This work presents insight beyond the above by exploring the potential for choosing much smaller blocks of weights as the units of clustering. We demonstrate the potential for 4- to 8-bit clustering with almost no accuracy loss. As a post-training quantization method, our work can be applied to off-the-shelf models without additional training and/or incorporated into a larger pipeline for further compression. Unlike other works such as in~\cite{han2016deep} require separate consideration of different layer types whereas LegoNet treats \textbf{all weights agnostic of architecture}. Further, our work does not just look at the two dimensional convolutional kernel, %~\cite{bell2015learning}, 
as is the case with previous clustering methods, we cluster \textbf{all dimensions of the convolutional layers} as seen in Fig.~\ref{fig:legonet}. This provides a much greater compression rate, as will be explored in the next sections. 
A further difference between LegoNet and previous works is the \textbf{lack of training needed to compress} the model, unlike other works. This is elaborated on in Section~\ref{related_work} as well as further justified in Section~\ref{methods}. Another departure from previous works is its \textbf{unique block shape (Legos)} that, similar to convolution filters, creates context that allows for accuracy preserving compression. This Lego shape is the crux of how our work \textbf{out preforms by a wide margin} all other works in this space. 
\end{comment}
The organization of this paper is as follows. We begin by motivating the problem addressed by LegoNet and outline its formulation in Section~\ref{motivation}. Then, in Section~\ref{related_work}, we survey related works. Next, in Section~\ref{methods}, we introduce the algorithms and techniques which comprise LegoNet. We apply LegoNet to various models and datasets and discuss our findings in the Section~\ref{results}. Finally, we put these results into context in Section~\ref{discussion} and explore their plausible implications in Section~\ref{conclusion}.

\section{Motivation}
\label{motivation}

%In this section, we provide the motivating problems that guide this work, and present a mathematical framework with which to view solutions through.

LegoNet is designed to address the need to use large high-capacity models that are trained and put them onto resource-constrained devices. An example for why one would want to do this would be to train a large network to a high degree of accuracy on a larger, more capable computer and then distribute it as a part of an application to a smaller device like a smartphone. %The main concern for designers when creating neural network based applications on these devices is their memory. 

%\begin{wrapfigure}{r}{.5\textwidth}
\begin{figure}
    \centering
    \includegraphics[width=.4\textwidth]{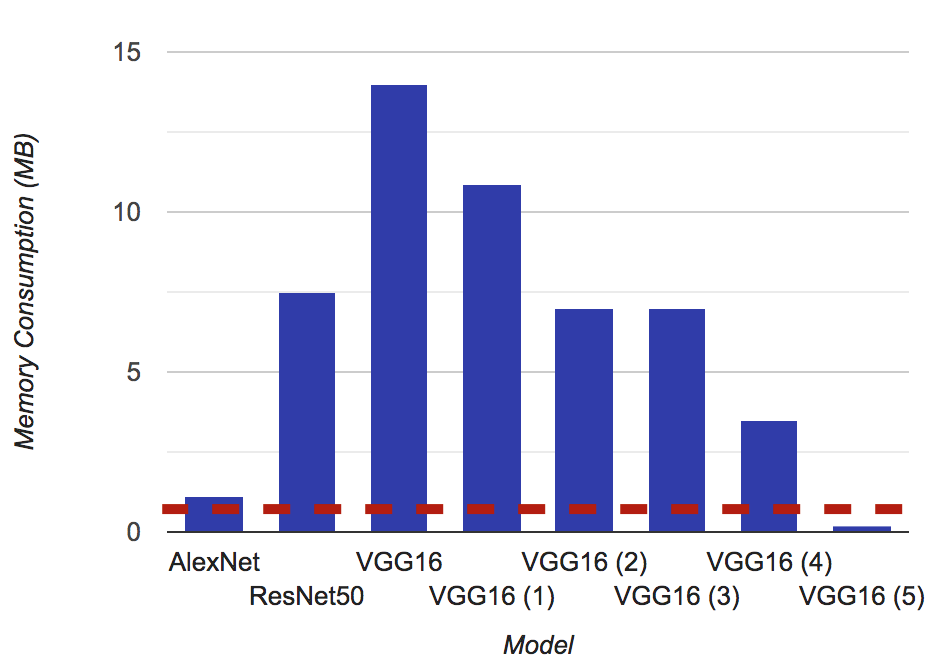}
    \caption{A comparison of memory consumption (blue bars) and the size of memory available in the STM32F7 micro-controller (red dotted line). VGG16 is compared to compressed versions, such as (1) Basis~\cite{Li_2019_ICCV}, (2) MBA~\cite{DBLP:journals/corr/abs-2006-03669}, (3) CRELU~\cite{DBLP:journals/corr/abs-2006-03669}, (4) CIRC~\cite{DBLP:journals/corr/abs-2006-03669} and our method, LegoNet (5). \textbf{Even though others may reduce other attributes (runtime, FLOPs, etc.) only LegoNet gets the model small enough to fit into the main memory.} }
    %\vspace{-2em}
    \label{fig:memory}
\end{figure}
%\end{wrapfigure}
As can be seen in Fig.~\ref{fig:memory}, for many of the most popular trained models like VGG-16 %~\cite{DBLP:journals/corr/SimonyanZ14a} 
and ResNet-50, %~\cite{DBLP:journals/corr/HeZRS15}
the memory required to run them is considerably larger than the memory available on micro-controllers (such as STM32F7). In order to run such a large model, one must find a way to reduce the memory footprint of the model at inference time.

In mathematical terms, if we have a deep neural network model $M$ approximating a function $f$ over a dataset $X$, then the goal of a compression method is to create a representation of $M$, $M'$ that requires a much smaller memory footprint. 
\begin{comment}
If we define the problem of training a model with regards to accuracy as 
\begin{equation}
\min_{M}dist(M[X], f[X]),
\end{equation}
where \textit{dist} represents the distance function from the real distribution (denoted by $f[X]$) and the distribution approximated by the model (denoted $M[X]$), then the problem of model compression could be formulated as 
\end{comment}

\begin{equation}
\min_{dist(M'[X], M[X]) < \epsilon} |M'|,
\end{equation}
where $dist$ is the distance between the outputs (the error), $\epsilon$ is some error tolerance, $M'$ is the compressed representation of the model, and $|M'|$ is the size of the memory footprint of $M'$. This is idealized version of the problem; in reality, we often concern ourselves with $dist(M'[X], M[X])$, which would be ensuring that the compressed model achieves as high an accuracy as the original model.

Ideally, $dist(M'[X], M[X]) = 0$. This would correspond to a lossless compression method. This is what some of the previous work (see Section~\ref{related_work}) as well as our work LegoNet (see Section~\ref{results}) achieve in terms of the model accuracy. However, it should be noted that there is merit in looking at what happens $0 < dist(M'[X], M[X]) < \epsilon$ where the error $\epsilon$ is small. %We will discuss this further in Section~\ref{discussion}. 

Moving forward, we define the compression ratio, or compression rate to be 
\begin{equation}
CR = \frac{|M|}{|M'|},
\end{equation}
%where $|M'|$ denotes the memory requirement for the compressed model. 
The larger a CR value is for a given representation of a model, the smaller $M'$ is. This provides a common language for discussing to what degree a method provides $M'$ such that $|M'| \ll |M|$. This will serve as our main metric and goal of this work.

\section{Background and Related Work}
\label{related_work}

%We provide an outline of previous work done in the field of memory footprint reduction techniques. We look at why these solutions fail to address all aspects of the research problem and why LegoNet fills this role.

%\subsection{Parameter Pruning} 

\textbf{Parameter Pruning} The increase in prevalence of neural networks in energy-constrained environments has been accompanied with a variety of techniques for model compression. Parameter pruning, whether by removing individual connections or entire convolutional kernels, has seen much success in reducing network size without negatively affecting accuracy by a significant amount. 
Further, recent work has shown that pruning has a theoretical upper bound on the lossless compression ratio that it can achieve~\cite{tanaka2020pruning}, which is far lower than our methods lossless compression ratio as we will show in Section~\ref{results}.%section Results~\ref{results}. 
%add more

\textbf{Reduced Model Design} There exist model architectures, such as MobileNet~\cite{howard2017mobilenets} and EfficientNet~\cite{tan2020efficientnet} which were designed specifically to run on resource-constrained devices such as Internet of Things. They leverage few parameters from the onset instead of compressing a larger, higher capacity model. This aids with reducing the training time and execution time but they achieve only at best 8x compression in comparison to ResNet-50 for less accuracy. This is worse compared to our results, in both compression and accuracy, as can be seen in our results (Section~\ref{results}). Further, our method can be applied to these models, for an even greater compression.

\textbf{Quantization} Parameter quantization reduces the number of bits required to store each weight.%~\cite{krishnamoorthi2018quantizing} 
These two methods are orthogonal and typically combined within a sequential compression pipeline as in~\cite{han2016deep}. Doing this, however, reduces the fundamental capacity %~\cite{DBLP:journals/corr/abs-2010-15327} 
of the network. Further, the compression ratio achieved by these methods is limited to at best the  ``word length'', or the bits needed to represent one value of the neural network, which is far lower than what we achieve in Section~\ref{results}.
%add more

\textbf{Minimal Clustering and Weight Representations} In the most extreme case, individual weights have been reduced to binary or ternary indices~\cite{ lee2020flexor, li2016ternary}. Weights preserving full floating-point or integer precision are clustered to 4- and 8-bit indices accompanied by accuracy loss in the range of a few percent~\cite{choukroun2019low}. %A similar method known as convolutional clustering~\cite{Son_2018_ECCV,DBLP:journals/corr/WuLWHC15}, consists of clustering the convolutional filters of a CNN. These methods only cluster the filters, and non of the weights. While this reduces the computation, it does not reduce the memory footprint hardly at all in comparison to LegoNet.

%\subsection{Post-Training Quantization} 

\textbf{Post-Training Quantization} Efforts in neural network quantization generally fall into one of two categories, quantization-aware training (QAT) and post-training quantization (PTQ).%~\cite{krishnamoorthi2018quantizing}
QAT methods require training or fine-tuning the model with the effect of learning quantized weights. This has been implemented with a quantization layer accompanied by a tailored gradient~\cite{esser2019learned}. % or alternatively with soft weight-sharing~\cite{ullrich2017soft}. 
In contrast, PTQ operates on an already accurate model and aims to quantize weights with minimal accuracy loss (possibly reduced further by fine-tuning). PTQ has been incorporated into compression pipelines typically accompanied by pruning and fine-tuning such as in~\cite{han2016deep}. %LegoNet is an example of PTQ, but with its \textbf{architecture agnostic} methods and use of weight blocks, it differs from the rest of these methods and achieves \textbf{higher compression ratios than all other current methods}~\ref{tab:cr_comparisons}.

%\subsection{Vector Quantization} 

\textbf{Vector Quantization} In the simplest sense, quantization entails clustering individual weights and replacing each with their cluster index~\cite{han2016deep}. Product and vector quantization methods consider columns of the weight matrix to be the unit of clustering allowing multiple weights to share a cluster index~\cite{stock2019and}. Alternative levels of granularity researched include kernel clustering \cite{lou2019autoq}. %and group learning \cite{ullrich2017soft}. These methods are similar to our solution, LegoNet. However our method \textbf{does not require training the clusters or the weights in anyway to achieve a higher compression ratio, while maintaining the same accuracy}. This allows for trained models to be compressed \textbf{without the need for additional data or training phases}. This also takes the form of clustering the convolutional filters of a CNN, as seen in \cite{pmlr-v80-wu18h}. Since this method limits itself to only convolutional filters, and not the rest of the model, it \textbf{does not achieve as high of a CR as LegoNet}. Further, many of these methods require quantization to occur while training, meaning they could not be used with off-the-shelf trained models.

%\subsection{Minimal Clustering and Weight Representations}

\section{Methods}
\label{methods}

The inspiration behind this work was originally to find a way to keep common or similar values found within a deep neural network in the cache of the computer during the inference. To that end, we looked at clustering \textit{blocks} of weights and keeping the centroids of these clusters in the cache for quick access. After noticing that with only a few more clusters needed to replace the full model without any loss to accuracy, we began testing on larger models. 
\subsection{Design}
Algorithm~\ref{alg:fine-pruning} shows the details of LegoNet. The algorithm begins with inputting the model $M$ the user wishes to compress, $K$ which is the number of clusters (and thus the Legos) that will be used, $b$ which is the dimension of the Lego pieces, i.e.,  each Lego will be $b \times b$ values. The parameter $b$ should be selected as the greatest common divisor of the dimensions of the layers, or some factor thereof, as will be shown later in Section~\ref{results}.

\begin{algorithm}
    \caption{LegoNet algorithm}
    \label{alg:fine-pruning}
    \begin{flushleft}
    \hspace{.25cm} \textbf{Input:} Model $M$, Int $K$, Int $b$ \\
    \hspace{.25cm} \textbf{Output:}  Legos[] $Indices$, Legos[] $l$\\
    \hspace{.25cm} Legos[] $blocks \leftarrow \emptyset $ \\
    \hspace{.25cm} $\forall$ Layers $l \in M$: \\
    \hspace{.75cm} $blocks \leftarrow blocks \cup $ Breakup($l$) \\
    \hspace{.25cm} $legos \leftarrow $ centers of clusters($blocks$) \\
    \hspace{.25cm} $Indices \leftarrow \emptyset$ \\
    \hspace{.25cm} $\forall b \in blocks$: \\
    \hspace{.75cm} $\forall l \in legos$: \\
    \hspace{1cm} if $distance(b,l)$ is $\min \forall b$: \\
    \hspace{1.25cm} $Indices \leftarrow Indices \cup ($index($b$), $l$)
    \end{flushleft}
    %\vspace{-1em}
\end{algorithm}

Next, the weights of the model are split into blocks of size $b \times b$ and stored into a data structure. Note that this is done completely \textbf{agnostic of what type of layer} the weights are a part of or where in the model they lie, a key difference from previous works. This is done without needing to take any consideration as to what layer or what kind of layer the weights come from. This agnosticism to where the weights are from is one differentiation between this work and prior works that were explained in Section~\ref{related_work}.

The blocks are clustered, and the centers of the resulting clusters are then recorded. These are the Legos we will be using to rebuild the model. For every block in the model, all of the centers are compared to it to see which is the minimal distance. The distance used in this work is Euclidean.%~\cite{gower1985properties} If the block is within this threshold from the Lego, the index of that Lego is recorded in the place of the block. This is repeated for all blocks in the model. %add more here 

%\begin{wrapfigure}{t}{0.5\textwidth}
\begin{comment} %cut for short paper

\begin{figure}

    %\vspace{-3em}
    \centering
    \includegraphics[width=.4\textwidth]{images/K plots/b=3.png}
    \caption{A figure showing the accuracy of the compressed model using Lego dimensions $b = 3$.}
    \label{fig:b_plot}
    %\vspace{-3em}
%\end{wrapfigure}
\end{figure}
\textbf{Why must $b$ evenly divide the layers} As will be seen in later experiments, if $b$ does not evenly divide the layer size as is the case of Fig.~\ref{fig:b_plot}, then some sort of compensation must be used. The use of padding throws off the algorithm, as all blocks that have padding will artificially be `closer' to one another than other since they will have the length of the layer being compressed $ \mod b \times $ the width of the layer values perfectly the same as they are all the padded value. This will cause them to cluster together, even though their non-padded values should be clustered elsewhere.

If truncation is used, that is to say the last block which does not fill an entire $b \times b$ remains its size, then these values could only be clustered with one another. Further, they would need an extra bit to differentiate the remainder blocks and the regular blocks. This would decrease the the compression efficiency and does not lead to higher accuracy in practice. Additionally, this would require extra logic to check what shape of block is being used, increasing the run time of the algorithm.

\end{comment}

\subsection{Theoretical Analysis}
In theory, a model's size would be given by $|M| = P*\textrm{wordlength}$, where $P$ is the number of parameters in the model and wordlength is the number of bits needed to represent the elements in the network. This would be because every weight contributes 1 wordlength worth of bits. For example a normal Double implementation of a network would have $\textrm{wordlength} = 64$. 

If we now look at our solution, for every parameter in the original model ($M$), in our compressed representation ($M'$), it would only contribute 
\begin{equation}\frac{\left\lceil\log_2 K\right\rceil}{b \times b}\end{equation} 
This means that the size of our compressed model is 

\begin{equation}|M'| = P\times \frac{\left\lceil\log_2 K\right\rceil}{b \times b} + b^2\times K\times \textrm{wordlength}\end{equation}
This comes from each weight being clustered in groups of $b \times b$ and then the index being recorded using $\left\lceil\log_2 K\right\rceil$ bits. The $b^2\times K\times \textrm{wordlength}$ comes from the code book of the Legos. Note that this term is negligible in most cases since for all models we looked at $K = 50, b = 4, \textrm{wordlength} = 32$ was sufficient for lossless compression (see Section~\ref{results}), which amounts to only \textbf{2048 bytes}. As such the size of the code book will be ignored for the rest of this analysis. 

Therefore, the theoretical compression ratio that would result from this method would be given by 

\begin{equation}\label{eq:cratio}
\textrm{CR} = \frac{|M|}{|M'|} = \frac{P*\textrm{wordlength}}{P*\frac{\left\lceil\log_2 K\right\rceil}{b \times b}} = \frac{b \times b \times \textrm{wordlength}}{\left\lceil\log_2 K\right\rceil},
\end{equation}

Note the $b \times b$ in the numerator and $K$, these are \textbf{the crux of what gives this work such a high compression ratio}. While many weight clustering methods from Section~\ref{related_work} only work on the single weight level, which would be the same as our method with $b = 1$, our method uses a higher $b$ value, which \textbf{increases the compression ratio quadratically}. Further, our method uses relatively small $K$'s, which has an inverse relation to the CR. With both of these factors, as will be seen in Section~\ref{results}, our method achieves a \textbf{64x lossless CR} and \textbf{128x CR with minimal loss}.

This equation is what we use through the rest of the paper to determine the CR value of a particular experiment. This value is tested on multiple datasets and models, as well as compared to other existing methods, in the next section.

\section{Results}
\label{results}

\newcommand\x{.15}

\begin{figure*}[h]
    %\centering
    \begin{flushleft}
    \subfloat{\includegraphics[width=\x\textwidth]{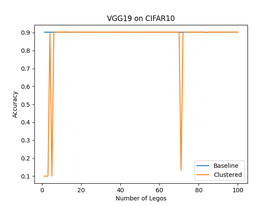}}
    \subfloat{\includegraphics[width=\x\textwidth]{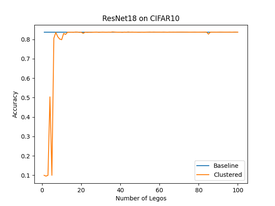}}
    \subfloat{\includegraphics[width=\x\textwidth]{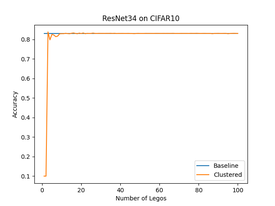}}
    \subfloat{\includegraphics[width=\x\textwidth]{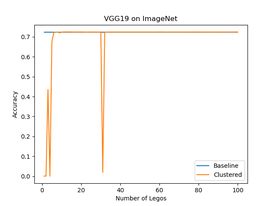}}
    \subfloat{\includegraphics[width=\x\textwidth]{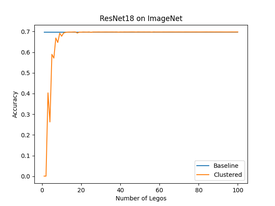}}
    \subfloat{\includegraphics[width=\x\textwidth]{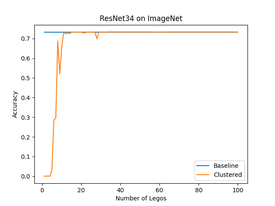}}
    \\ %b = 2
    \subfloat{\includegraphics[width=\x\textwidth]{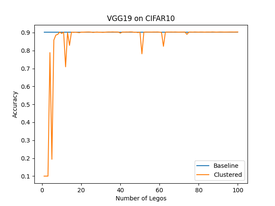}}
    \subfloat{\includegraphics[width=\x\textwidth]{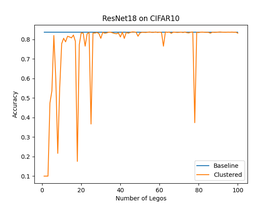}}
    \subfloat{\includegraphics[width=\x\textwidth]{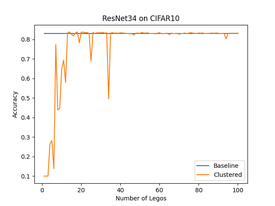}}
    \subfloat{\includegraphics[width=\x\textwidth]{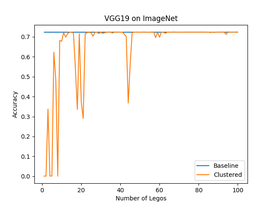}}
    \subfloat{\includegraphics[width=\x\textwidth]{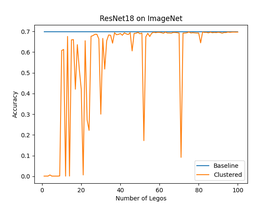}}
    \subfloat{\includegraphics[width=\x\textwidth]{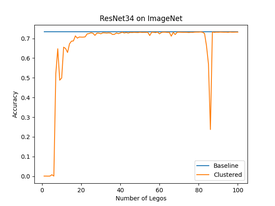}}
     \\ %b = 4
    \subfloat{\includegraphics[width=\x\textwidth]{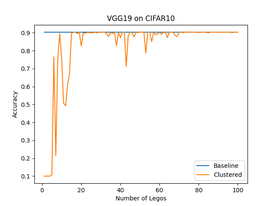}}
    \subfloat{\includegraphics[width=\x\textwidth]{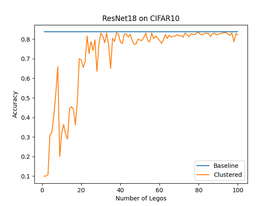}}
    \subfloat{\includegraphics[width=\x\textwidth]{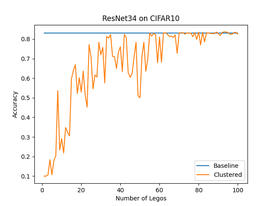}}
    \subfloat{\includegraphics[width=\x\textwidth]{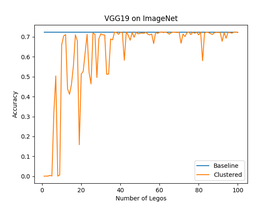}}
    \subfloat{\includegraphics[width=\x\textwidth]{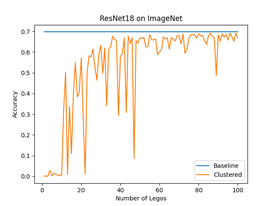}}
    \subfloat{\includegraphics[width=\x\textwidth]{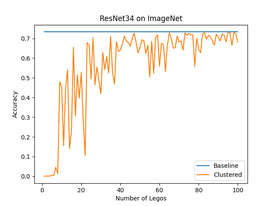}}
    \caption{Analysis of $K$ values for a given $b$ value, $M$ and dataset. Note, $b = 4$ is the best for almost all models used in terms of compression ratio for a given accuracy tolerance.}
    \label{fig:b_values}
\end{flushleft}       
%\vspace{-3.5em}
\end{figure*}

%We present the results of empirically applying LegoNet to a variety of models and datasets.

\subsection{Setup}

%\subsubsection{Models and Datasets} % ResNet-50, VGG16, CIFAR10, ImageNet

\noindent\textbf{Models and Datasets} We evaluate our methods on pre-trained instances of VGG-16 and VGG-19%~\cite{DBLP:journals/corr/SimonyanZ14a} 
and ResNet-18, -34, and -50.%~\cite{DBLP:journals/corr/HeZRS15}. 
In order to demonstrate consistency for both small- and large-scale classification tasks, we apply these networks to the CIFAR10 and ImageNet datasets. The baseline accuracies for each model/database pair are described in Table~\ref{tab:my_label}. 

\begin{table}[h]
    \centering
    \small
    \caption{Top-1 accuracies for each model and dataset before and after applying LegoNet. LegoNet accuracies are reported for the best $K$ value with regards to accuracy as well as the lowest $K$ value with maximal accuracy.}
    \begin{tabular}{llllll}
    %add new values from "sup"
        Model & Dataset & Baseline & LegoNet-A & LegoNet-C \\
        \hline \hline
        VGG16 & CIFAR10 & 90.02 & 90.07 & 87.62 \\ % 90.07 \\
        \hline
        VGG16 & ImageNet & 71.59 & 71.57  & 68.96 \\ % 71.59 \\
        \hline
        VGG19 & CIFAR10 & 90.03 & 90.44  & 89.60  \\
        \hline
        VGG19 & ImageNet & 72.38 & 72.45 & 70.39 \\
        \hline
        ResNet-18 & CIFAR10 & 83.70 & 83.44 & 81.61 \\ 
        \hline
        ResNet-18 & ImageNet & 69.76 & 69.21 & 67.66  \\ 
        \hline
        ResNet-34 & CIFAR10 & 83.08 & 83.63 & 81.34 \\ 
        \hline
        ResNet-34 & ImageNet & 73.31 & 72.91 & 71.03 \\ 
        \hline
        ResNet-50 & CIFAR10 & 85.42 & 85.28 & 83.76 \\ % 85.28 \\
        \hline
        ResNet-50 & ImageNet & 76.13 & 75.78 &  73.91 \\ % 75.78 \\
        \hline
        \hline
    \end{tabular}
    %\vspace{-2.5em}
\label{tab:my_label}
\end{table}

\textbf{Hyperparameters} There are a few hyperparameters that are used to control the compression ratio and accuracy of our method. Our experiments reveal insight into the effects of the following parameters on LegoNet's accuracy:
%\begin{itemize}
    \par - The number $K$ of cluster centers used as Legos.
    %\item The maximum distance $\delta$ from each cluster centroid to each block replaced by that centroid.
    %\par - The network layers $M$ chosen to participate in clustering. While we experimented with restricting the layers used, the best results were from compressing all layers.  
    \par - The size of the Legos, $b$.
    %\item The clustering algorithm utilized.
%\end{itemize}

To this end, we experiment exhaustively across the range of possible values for $K$. We further experimented with various values for $b$. In Fig.~\ref{fig:b_values}, we looked at variety of models with block sizes of $b=1,2,4$; these values evenly divide all dimensions of the layers of interest. In practice, for $b$ to evenly divide all layers that are being compressed, and because the larger $b$ leads to higher compression, we suggest that $b$ be chosen as the greatest common divisor of the layers, which for both ResNet-50 and VGG-16 is 4. We also vary the layers clustered between convolutional layers only, linear layers only, and both convolutional and linear layers in order to show that LegoNet is layer agnostic in terms of accuracy preservation. 

The K-means clustering algorithm was used for all of the experiments. %This choice introduces some element of randomness in the outcome of a particular run of LegoNet based on the random initialization for the cluster heads. This is what causes the noise seen throughout the results. We repeated all experiments and found this noise present in at least some of the $K$ values for all runs. This can be overcome by running the algorithm with multiple seeds to find the best values.

\textbf{Metrics} We are concerned with the maximum compression ratio that can be realized without significant accuracy loss. We determine this by recording the top-1 classification accuracies for each model, dataset, and parameter configuration after block-level weight clustering. Considering our choice of block size, the total compression rate we achieve for each experiment is given by $\frac{4^2\times 32}{\log_2 k}$, where 32 comes from the fact that floating point values used in PyTorch have a word length of 32. If double precision is used instead, then the compression rate would be $\frac{4^2\times 64}{\log_2 k}$. 
\begin{comment}
We also concern ourselves with the accuracy of the model. Of course, lossless compression (i.e., compression with no loss to accuracy) with the highest compression ratio is the goal. However, if a small amount of accuracy could be sacrificed for a great increase to the compression ratio, then this should be considered. This trade off is explored below.
\end{comment}

\subsection{Experimental Results}

%\subsubsection{Compression Ratio}
\textbf{Compression Ratio} 
Due to allowing 16 weights (each block of $4\times4$ weights) to share a single index, our method allows for \textbf{64x compression rate with no accuracy loss} when quantizing to 8 bits (the number of bits needed to represent the legos uniquely) across both convolutional and linear layers, as given by Equation~\ref{eq:cratio}. For LegoNet-C, which has a smaller $K$ value but allows for a small accuracy loss, we achieve almost double this rate. Table~\ref{tab:cr_comparisons} compares our compression rate against several recent and state-of-the-art similar compression methods. LegoNet shows significant improvement over pure quantization methods and even beats pipelines including both pruning and quantization.

\begin{figure*}[t]
    
    \centering
    \subfloat{\includegraphics[width=0.23\textwidth]{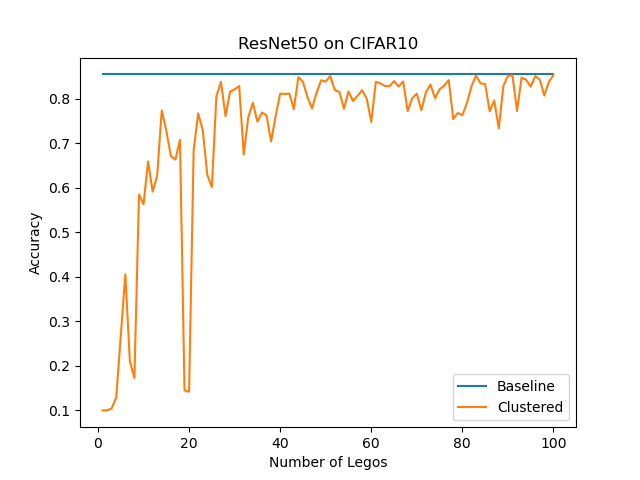}}
    \subfloat{\includegraphics[width=0.23\textwidth]{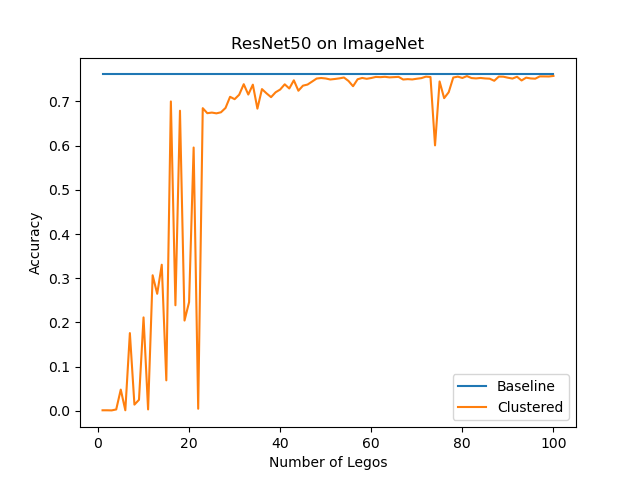}}
    %\\
    \subfloat{\includegraphics[width=0.23\textwidth]{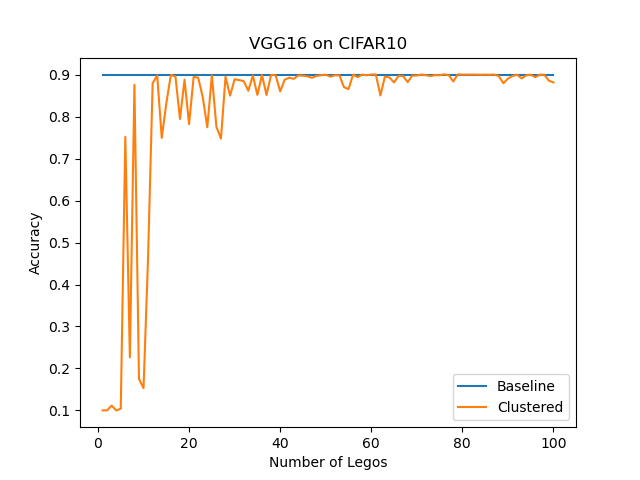}}
    \subfloat{\includegraphics[width=0.23\textwidth]{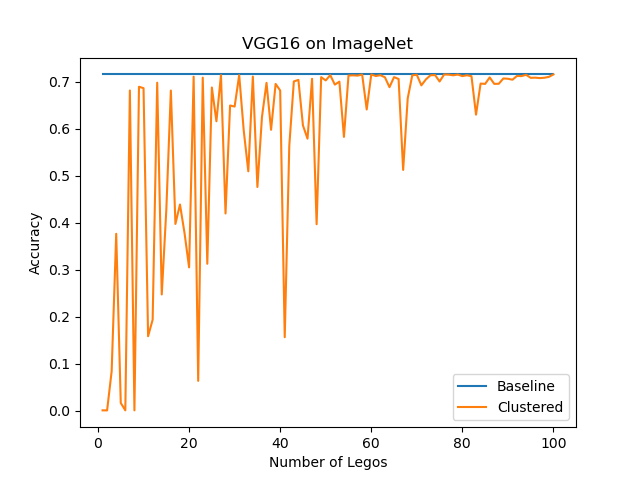}}
    
    \caption{Accuracy for a variety of $K$ values (orange line) compared to the baseline (blue line).}
    \label{fig:k_plots}
    
\end{figure*}

%\noindent\textbf{Number of legos for no accuracy loss during compression.} 
\textbf{LegoNet-Accuracy}
The number of Legos $K$ used to cluster a network is the predominant hyperparameter, along with $b$, in achieving a high compression rate. Whereas $b$ is something that is chosen based on the architecture, $K$ can be varied at will. Choosing the correct value, therefore, requires careful consideration. Fig.~\ref{fig:k_plots} shows empirically that some networks can retain peak accuracy with as few as $K=20$ Legos without any loss of accuracy. In all cases, \textbf{original accuracy is assured with at least $K\leq50$ Legos}. We tested this on VGG, ResNet, and LeNet networks with identical results. While accuracy trends downward as the number of clusters is reduced, we note that there is a large amount of variance and chaos in the accuracy. For example, there are outlying small values of $K$ (i.e., number of Legos) which can preserve the original accuracy. Analysis of a specific network could reveal an acceptable value as low as $K=8$. In general, letting $K <= 64$ allows for all weights to be indexed with a single byte with no loss of model accuracy.

\textbf{LegoNet-Compression} As shown in Fig.~\ref{fig:k_plots}, for some surprisingly low $K$ values, the Legos can represent the model with a higher compression ratio for negligible loss to accuracy. In order to find this value, we iteratively increase $K$ until a desired error tolerance is met. Although reclustering for every new $K$ may seem costly, since this is not to be done on device, the memory footprint of this operation need not be considered. Further, we have found that these values exist at considerably lower values than the lossless $K$ values across all datasets and models that we have tested. For ResNet-50 on the ImageNet dataset, this $K$ value allows for \textbf{half the bits} to be used than the lossless value for \textbf{only $<3\%$ loss to accuracy}. Note this is less than the accuracy loss found in the state-of-the-art~\cite{esser2019learned}, while we additionally have a 2.6X better compression ratio. 

\begin{table}[t]
    \centering
    \caption{Compression Rate (CR) comparison versus other compression methods using the Imagenet dataset on the ResNet-50 model. All values that are not our results were taken from the paper directly.}
    \begin{tabular}{lll}
        Method & CR & Accuracy Loss \\
        \hline \hline
        \textbf{LegoNet-A} & \textbf{64} & 0\% \\
        \hline
        \textbf{LegoNet-C} & \textbf{128} & 2.8\% \\
        \hline
        MMSE~\cite{choukroun2019low} & 8 & 1.7\%\\ %Integer Quantization
        \hline
        DC~\cite{han2016deep} & 49 & 0\% \\
        \hline
        LSSQ~\cite{esser2019learned} & 49 & 3.2\% \\
        \hline
        %RQ ST \cite{louizos2018relaxed} & 16 \\
        ARQ~\cite{stock2019and} & 26 & 0\%\\
        \hline
        VQ~\cite{gong2014compressing} & 24 & $<$ 1\% \\
        \hline
        DKM\footnote{With Retraining} \cite{pmlr-v80-wu18h} & 16 & 1.3\% \\
        \hline
        PQFT~\cite{Martinez_2021_CVPR} & 19 & 1.57\% \\
        \hline
        DMC~\cite{Gao_2020_CVPR} & 2.22 & 0.8\% \\
        %\hline
        %GAL~\cite{DBLP:journals/corr/abs-1903-09291} & 2.22 & 4.35\%
        %\\
        \hline
        AN~\cite{DBLP:journals/corr/abs-2006-05624} & 3.8 & 0.67\% \\
        \hline
        CC~\cite{Li_2021_CVPR} & 2.07 & 0.56\%
        \\
        \hline
        CHIP~\cite{DBLP:journals/corr/abs-2110-13981} & 2.26 & 0.89\%
        \\
        \hline
        \hline
        %\vspace{-3em}
    \end{tabular}
    \label{tab:cr_comparisons}
\end{table}

\section{Discussion}
\label{discussion}
%In this section, we discuss the benefits and use cases of this work, and its benefits outside of our initial goals.

\textbf{Why LegoNet Works} We can compare LegoNet to other state-of-the-art methods in recent literature theoretically based on Equation~\ref{eq:cratio}. Methods, such as pruning, knowledge distillation, and reduced model design, work by reducing the number of parameters in the model. In our equation, this would correspond to making $P$ smaller. Quantization, on the other hand, aims to reduce the number of bits used to represent a particular set or all of the parameters of the model. This would be the same as reducing $\textrm{wordlength}$. Methods such as weight sharing and other clustering techniques work to use a lookup table to replace some values in the model with a reduced representation, which would be the same as the denominator in the equation, however in this case $b \times b = 1 \times 1$, since they cluster single values.

As can be seen, the effect of aforementioned methods in the overall compression ratio is a linear or sub-linear relation. This is why LegoNet achieves such a high compression ratio when compared to other methods because it uses the b value, which has a \textbf{quadratic effect on compression}. Further, LegoNet leverages the $\frac{1}{\left\lceil\log_2 K\right\rceil}$ on top of the $b \times b$. These factors are why LegoNet achieves such high compression ratios and the crux of this work. 

Another factor in LegoNet's superiority in comparison to vector quantization methods, even further than just the consideration of all layers instead of just convolutional layers, is the Lego shape. Instead of vectors, which are 1-dimensional, LegoNet uses 2-dimensional blocks. Similar to how convolutional kernels provide context using a block shape, so too does the Lego shape. This departure from previous works allows us to have \textbf{higher accuracy while requiring fewer entries in our codebook}. This use of shape is instrumental to the \textbf{vastly higher compression ratio}.

\textbf{Updating Lego-ed Model} As mentioned earlier, LegoNet works on trained models. This is by design, because the use case of this work is to compress premade models. However LegoNet will not always increase the accuracy, so if the goal is to fine-tune a model while it is being compressed, LegoNet will not do this. However, if new data is acquired and the model needs to be fine-tuned, the Lego-fied model can be rebuilt into a normal model, fine-tuned using normal backpropagation, and then LegoNet algorithm can be applied to the model. 
\begin{comment}
\textbf{Further Compression} LegoNet's core objective is complementary to compression methods. It can be coupled with others compression methods such as~\cite{han2016deep}. If the capacity of the model is not of concern, the user can utilize structural pruning to reduce the complexity of the model before applying LegoNet, the compression ratio would be higher; however, this work focuses on non-destructive methods. Additionally, if unstructured pruning is utilized, we hypothesize that the induced zeros will reduce the dimensionality of the Legos. This would potentially make clustering even more efficient and reduce the number of Legos needed to achieve lossless compression.

\textbf{Hardware Consideration} For optimal compression, it is obvious that the smallest $K$ value should be used. However, for compressing the 64-bit floating points down to an index in the dictionary, conventional processors have a fixed word length. This means that even if $K$ is 8, which would only require theoretically 3 bits to represent it, the practical implementation would use 4 bits. This can be rectified by using special custom-designed hardware that has the ability to use fewer number bits. This will increase the compression ratio and reduce the memory footprint further.
\end{comment}

\section{Conclusion}
\label{conclusion}

We introduced LegoNet, a block-based weight compressing method to reduce a deep neural networks memory footprint. We showed how it addresses the issue of fitting a large trained network, like VGG-16 or ResNet-50, onto an embedded or low-memory device. We showed that it outperforms other methods such as Deep Compression, Deep $k$-Means, and Vector Quatnitzation in this space while not needing any data to achieve a compression ratio of 64x with LegoNet-A, 128x with LegoNet-C, and could be used for higher compression ratios with no or minimal loss to accuracy on ResNet-50 testing on the ImageNet 2012 dataset. We explained LegoNet's weight agnostic design and block-based method differs from other, previous works allows it to achieve such high compression ratios.

\bibliographystyle{splncs04}
\bibliography{egbib}

@String(CVPR= {IEEE Conf. Comput. Vis. Pattern Recog.})

@String(ICCV= {Int. Conf. Comput. Vis.})

@String(CVPR  = {CVPR})

@String(ICCV  = {ICCV})

@article{liu2017survey,
  title={A survey of deep neural network architectures and their applications},
  author={Liu, Weibo and Wang, Zidong and Liu, Xiaohui and Zeng, Nianyin and Liu, Yurong and Alsaadi, Fuad E},
  journal={Neurocomputing},
  volume={234},
  pages={11--26},
  year={2017},
  publisher={Elsevier}
}

@misc{han2016deep,
      title={Deep Compression: Compressing Deep Neural Networks with Pruning, Trained Quantization and Huffman Coding}, 
      author={Song Han and Huizi Mao and William J. Dally},
      year={2016},
      eprint={1510.00149},
      archivePrefix={arXiv},
      primaryClass={cs.CV}
}

@misc{nguyen2021wide,
      title={Do Wide and Deep Networks Learn the Same Things? Uncovering How Neural Network Representations Vary with Width and Depth}, 
      author={Thao Nguyen and Maithra Raghu and Simon Kornblith},
      year={2021},
      eprint={2010.15327},
      archivePrefix={arXiv},
      primaryClass={cs.LG}
}

@misc{krishnan2019structural,
      title={Structural Pruning in Deep Neural Networks: A Small-World Approach}, 
      author={Gokul Krishnan and Xiaocong Du and Yu Cao},
      year={2019},
      eprint={1911.04453},
      archivePrefix={arXiv},
      primaryClass={cs.LG}
}

@article{Gou_2021,
   title={Knowledge Distillation: A Survey},
   ISSN={1573-1405},
   url={http://dx.doi.org/10.1007/s11263-021-01453-z},
   DOI={10.1007/s11263-021-01453-z},
   journal={International Journal of Computer Vision},
   publisher={Springer Science and Business Media LLC},
   author={Gou, Jianping and Yu, Baosheng and Maybank, Stephen J. and Tao, Dacheng},
   year={2021},
   month={Mar}
}

@InProceedings{pmlr-v80-wu18h,
  title = 	 {Deep k-Means: Re-Training and Parameter Sharing with Harder Cluster Assignments for Compressing Deep Convolutions},
  author =       {Wu, Junru and Wang, Yue and Wu, Zhenyu and Wang, Zhangyang and Veeraraghavan, Ashok and Lin, Yingyan},
  booktitle = 	 {Proceedings of the 35th International Conference on Machine Learning},
  pages = 	 {5363--5372},
  year = 	 {2018},
  editor = 	 {Dy, Jennifer and Krause, Andreas},
  volume = 	 {80},
  series = 	 {Proceedings of Machine Learning Research},
  month = 	 {10--15 Jul},
  publisher =    {PMLR},
  url = 	 {http://proceedings.mlr.press/v80/wu18h.html}
}

@article{esser2019learned,
  title={Learned step size quantization},
  author={Esser, Steven K and McKinstry, Jeffrey L and Bablani, Deepika and Appuswamy, Rathinakumar and Modha, Dharmendra S},
  journal={arXiv preprint arXiv:1902.08153},
  year={2019}
}

@article{stock2019and,
  title={And the bit goes down: Revisiting the quantization of neural networks},
  author={Stock, Pierre and Joulin, Armand and Gribonval, R{\'e}mi and Graham, Benjamin and J{\'e}gou, Herv{\'e}},
  journal={arXiv preprint arXiv:1907.05686},
  year={2019}
}

@article{lou2019autoq,
  title={Autoq: Automated kernel-wise neural network quantization},
  author={Lou, Qian and Guo, Feng and Liu, Lantao and Kim, Minje and Jiang, Lei},
  journal={arXiv preprint arXiv:1902.05690},
  year={2019}
}

@article{lee2020flexor,
  title={FleXOR: Trainable Fractional Quantization},
  author={Lee, Dongsoo and Kwon, Se Jung and Kim, Byeongwook and Jeon, Yongkweon and Park, Baeseong and Yun, Jeongin},
  journal={arXiv preprint arXiv:2009.04126},
  year={2020}
}

@article{li2016ternary,
  title={Ternary weight networks},
  author={Li, Fengfu and Zhang, Bo and Liu, Bin},
  journal={arXiv preprint arXiv:1605.04711},
  year={2016}
}

@inproceedings{choukroun2019low,
  title={Low-bit Quantization of Neural Networks for Efficient Inference.},
  author={Choukroun, Yoni and Kravchik, Eli and Yang, Fan and Kisilev, Pavel},
  booktitle={ICCV Workshops},
  pages={3009--3018},
  year={2019}
}

@article{gong2014compressing,
  title={Compressing deep convolutional networks using vector quantization},
  author={Gong, Yunchao and Liu, Liu and Yang, Ming and Bourdev, Lubomir},
  journal={arXiv preprint arXiv:1412.6115},
  year={2014}
}

@misc{tanaka2020pruning,
      title={Pruning neural networks without any data by iteratively conserving synaptic flow}, 
      author={Hidenori Tanaka and Daniel Kunin and Daniel L. K. Yamins and Surya Ganguli},
      year={2020},
      eprint={2006.05467},
      archivePrefix={arXiv},
      primaryClass={cs.LG}
}

@misc{howard2017mobilenets,
      title={MobileNets: Efficient Convolutional Neural Networks for Mobile Vision Applications}, 
      author={Andrew G. Howard and Menglong Zhu and Bo Chen and Dmitry Kalenichenko and Weijun Wang and Tobias Weyand and Marco Andreetto and Hartwig Adam},
      year={2017},
      eprint={1704.04861},
      archivePrefix={arXiv},
      primaryClass={cs.CV}
}

@misc{tan2020efficientnet,
      title={EfficientNet: Rethinking Model Scaling for Convolutional Neural Networks}, 
      author={Mingxing Tan and Quoc V. Le},
      year={2020},
      eprint={1905.11946},
      archivePrefix={arXiv},
      primaryClass={cs.LG}
}

@InProceedings{Martinez_2021_CVPR,
    author    = {Martinez, Julieta and Shewakramani, Jashan and Liu, Ting Wei and Barsan, Ioan Andrei and Zeng, Wenyuan and Urtasun, Raquel},
    title     = {Permute, Quantize, and Fine-Tune: Efficient Compression of Neural Networks},
    booktitle = {Proceedings of the IEEE/CVF Conference on Computer Vision and Pattern Recognition (CVPR)},
    month     = {June},
    year      = {2021},
    pages     = {15699-15708}
}

@InProceedings{Gao_2020_CVPR,
author = {Gao, Shangqian and Huang, Feihu and Pei, Jian and Huang, Heng},
title = {Discrete Model Compression With Resource Constraint for Deep Neural Networks},
booktitle = {IEEE/CVF Conference on Computer Vision and Pattern Recognition (CVPR)},
month = {June},
year = {2020}
}

@article{DBLP:journals/corr/abs-2006-05624,
  author    = {Utkarsh Nath and
               Shrinu Kushagra},
  title     = {Better Together: Resnet-50 accuracy with {\textdollar}13x{\textdollar}
               fewer parameters and at {\textdollar}3x{\textdollar} speed},
  journal   = {CoRR},
  volume    = {abs/2006.05624},
  year      = {2020},
  url       = {https://arxiv.org/abs/2006.05624},
  eprinttype = {arXiv},
  eprint    = {2006.05624},
  bibsource = {dblp computer science bibliography, https://dblp.org}
}

@InProceedings{Li_2021_CVPR,
    author    = {Li, Yuchao and Lin, Shaohui and Liu, Jianzhuang and Ye, Qixiang and Wang, Mengdi and Chao, Fei and Yang, Fan and Ma, Jincheng and Tian, Qi and Ji, Rongrong},
    title     = {Towards Compact CNNs via Collaborative Compression},
    booktitle = {Proceedings of the IEEE/CVF Conference on Computer Vision and Pattern Recognition (CVPR)},
    month     = {June},
    year      = {2021},
    pages     = {6438-6447}
}

@article{DBLP:journals/corr/abs-2110-13981,
  author    = {Yang Sui and
               Miao Yin and
               Yi Xie and
               Huy Phan and
               Saman A. Zonouz and
               Bo Yuan},
  title     = {{CHIP:} CHannel Independence-based Pruning for Compact Neural Networks},
  journal   = {CoRR},
  volume    = {abs/2110.13981},
  year      = {2021},
  url       = {https://arxiv.org/abs/2110.13981},
  eprinttype = {arXiv},
  eprint    = {2110.13981},
  bibsource = {dblp computer science bibliography, https://dblp.org}
}

@article{DBLP:journals/corr/abs-2006-03669,
  author    = {James O'Neill},
  title     = {An Overview of Neural Network Compression},
  journal   = {CoRR},
  volume    = {abs/2006.03669},
  year      = {2020},
  url       = {https://arxiv.org/abs/2006.03669},
  eprinttype = {arXiv},
  eprint    = {2006.03669},
  bibsource = {dblp computer science bibliography, https://dblp.org}
}

@InProceedings{Li_2019_ICCV,
author = {Li, Yawei and Gu, Shuhang and Gool, Luc Van and Timofte, Radu},
title = {Learning Filter Basis for Convolutional Neural Network Compression},
booktitle = {Proceedings of the IEEE/CVF International Conference on Computer Vision (ICCV)},
month = {October},
year = {2019}
}

\end{document}